%% file: PAI2026-GEMOR-GS.tex
\newif\ifPAIanonymous
\PAIanonymousfalse 

\documentclass[12pt]{article}
\usepackage{newtxtext,newtxmath}
\usepackage{graphicx}
\usepackage[letterpaper,margin=1in]{geometry}
\usepackage{booktabs}
\usepackage{multirow}
\usepackage{amsmath}
\usepackage{float} 

\renewenvironment{abstract}{\quotation}{\endquotation}
\date{}

\makeatletter
\renewcommand{\fnum@figure}{\textbf{Figure \thefigure}}
\renewcommand{\fnum@table}{\textbf{Table \thetable}}
\makeatother
\usepackage{PAIcite}
\usepackage{url}
\usepackage[hidelinks]{hyperref}
\hypersetup{pdfsubject={LiDAR-Inertial-Visual SLAM with 3D Gaussian Splatting},
pdfkeywords={3D Gaussian Splatting, LiDAR-camera calibration, SLAM, Global reconstruction, Relocalization, World model}}
\newcommand{\PAItitle}{GEMOR-GS: Global Extrinsic-calibration-refined Mapping and Online Relocalization for 3D Gaussian-Splatting SLAM} 
\title{\bfseries\boldmath\PAItitle\textsuperscript{1}}
\ifPAIanonymous
  \author{Anonymous submission}
\else
  \input{PAI_author_details.tex}

\fi
\AtBeginDocument{%
\ifPAIanonymous
\hypersetup{pdfauthor={Anonymous submission}}%
\else
\hypersetup{pdfauthor={Ming Ding, Yuchen Yang, Xinyu Wu, Wei Wang, Hesheng Wang}}%
\fi
}
\begin{document}
\maketitle
\ifPAIanonymous\else
\footnotetext[1]{Code and data are publicly available at \url{https://github.com/Vector-4D/gemor-gs}.}
\fi

\begin{abstract}\bfseries\boldmath
Photorealistic and metrically consistent world representations are a primary requirement for LiDAR--inertial--visual SLAM systems that serve robots over continuous operation.
3D Gaussian splatting makes such representations attainable because it supplies the explicit, renderable, and queryable scene model.
In this paper, we present a novel method, GEMOR-GS (Global Extrinsic-calibration-refined Mapping and Online Relocalization for 3D Gaussian-Splatting SLAM) to make the Gaussian map geometrically accurate, globally refined, and re-enterable.
The refined model can give the environment a better appearance for robot perception and decision-making, such as object detection, path planning, relocalization, and mapping. 
In order to improve the quality of the 3DGS map, before modeling, the LiDAR--camera extrinsic is estimated offline first by a direct-method LiDAR--camera alignment over dedicated calibration segments.
It can be reused by all following online and offline processes.
Under the calibrated geometry, keyframe poses and Gaussian map parameters are refined alternately under a gated differentiable-rendering and LiDAR--projection objective.
Then, the method rebuilds the final global reconstruction from the refined keyframe schema with all online constraints lifted.
For the robot SLAM lifecycle, this method can recover the pose from the 3DGS map after tracking loss by an XFeat bypass map, RANSAC-PnP coarse localization, Gaussian RGB-D warp refinement, followed by a global ICP-based geometric route to raise the accuracy of the pose.
Experiments on several different datasets (one self-collected and five public sequences) quantified each element.
Our offline reconstruction wins all paired keyframes against the online map by 14.3--19.9 dB.
Compared with another method (Gaussian-LIC2), our method can improve the $\mathrm{PSNR}$ by approximately 9 dB.
Held-out relocalization succeeds on 86--100\% of queries on most estimated sequences with reliability tied to extrinsic quality, and the ICP-based geometric route sustains 98--100\% success even where visual matching fails for the extreme weak-texture scene.
The experiment result shows that the proposed method can improve the geometric accuracy, appearance fidelity, and re-enterability of the 3DGS map.
It can be used as a dependable robot world model representation for robot SLAM and downstream tasks.
\end{abstract}

\section{Introduction}
\label{sec:intro}

\subsection{Background}
\label{sec:intro-bg}
In recent years, 3D Gaussian splatting (3DGS) has become the dominant explicit radiance-field representation for SLAM and dense reconstruction, and LiDAR--inertial--visual (LIV) tightly coupled SLAM systems augmented with it promise robots a photorealistic and metrically consistent world representation \cite{kerbl3Dgaussians,zheng2025fastlivo2,hong2025gslivo}.
Beyond rendering benchmarks, 3DGS models are also increasingly deployed as world-model representations:
embodied agents build Gaussian world representations of goal states for visual rearrangement \cite{melik24gaussianworld};
end-to-end driving stacks use splatting as the scene representation of a predictive world model \cite{yan2025renderworld};
and object-centric Gaussian world models learn action-conditioned rigid-body dynamics to support model predictive control \cite{kreber2026mrogwm}.
This trend assigns the SLAM-produced 3DGS map a more demanding role than that of an offline asset scored by rendering metrics.
The map must serve as a queryable and dependable world representation that supports localization, mapping, and downstream decision-making over continuous operation.
Geometric accuracy (calibration and poses), appearance fidelity (reconstruction quality), and re-enterability (relocalization) are first-order prerequisites.

However, most current LIV fusion pipelines are still not queryable and dependable enough for robot SLAM and world-model deployment.
\emph{Geometric accuracy} demands trustworthy spatiotemporal calibration.
LiDAR--camera extrinsic residuals and the nearest-pairing offset are written into every keyframe's depth--color binding, and hence into the Gaussian supervision.
The ghosting observed in fused maps mixes three coupled sources: spatial extrinsic residuals, the temporal pairing offset, and pose suboptimality.
The online filter produces only locally optimal Markov estimates of pose and absorbs extrinsic residuals only up to the observability limit.
A trustworthy calibration should be applied to ensure the accuracy of the extrinsic.
The extrinsic must be estimated before refining the poses and map.
And the propagation of extrinsic quality into reconstruction and relocalization must be quantified, not just be assumed.
\emph{Appearance fidelity} demands a global refinement stage.
The real-time budget admits only limited per-frame optimization over a field-of-view (FoV) sliding window, without densification and with low-order spherical harmonics.
Therefore, the online map remains a tracking aid rather than the final reconstruction.
The benefit of calibration and pose refinement must propagate into reconstruction quality, which requires a global reconstruction stage that inherits the refined calibration and poses.
\emph{Re-enterability} demands a recovery mechanism in the robot SLAM lifecycle.
Tracking loss is inevitable in real deployments, arising from fast rotation, motion blur, lighting change, and dynamic occlusion.
If the 3DGS map does not provide the ability to recover the pose, a robot cannot re-enter the map, resume mapping, or support downstream tasks.
Most existing 3DGS SLAM work stops at offline rendering evaluation.
For using 3DGS mapping in robot SLAM, the lifecycle of tracking, loss, recovery, and resumption would be an indispensable consideration.

\subsection{Objective}
\label{sec:intro-contrib}
In this research, we present a novel method --- GEMOR-GS (\textbf{G}lobal \textbf{E}xtrinsic-calibration-refined \textbf{M}apping and \textbf{O}nline \textbf{R}elocalization for 3D \textbf{G}aussian-Splatting \textbf{S}LAM), which is a LiDAR--inertial--visual SLAM system that estimates poses under a real-time budget and archives keyframes.
It upgrades the SLAM-produced 3DGS map from an offline render asset to a dependable robot world model.
It is a representation whose geometric accuracy, appearance fidelity, and re-enterability can be relied upon by localization, mapping, and downstream decision-making over continuous operation for a robot even when tracking loss occurs.
In this method, for the extrinsic calibration, a direct-method module is used to estimate the LiDAR--camera extrinsic offline over dedicated calibration segments and supplies it to other modules.
The offline track then refines keyframe poses and the Gaussian map alternately, purifying the depth--color supervision that drives Gaussian training.
For offline mapping, a global reconstruction stage re-initializes the Gaussian model from the refined keyframe schema and executes global densification, pruning, and spherical-harmonics order elevation, turning the online sliding-window map into the final high-fidelity reconstruction.
For relocalization, a bypass module conducts an image-based method (matching XFeat descriptors against a lightweight relocalization database, obtaining a coarse pose via RANSAC-PnP, and refining it via Gaussian RGB-D warp alignment) and a LiDAR-based method (using an ICP-based geometric route to register the query LiDAR scan against the point cloud of the reconstructed map), enabling the robot to recover its global pose after tracking loss.

This work makes the following three primary contributions:
\begin{enumerate}
\item \textbf{A global-optimized reconstruction stage.}
Our method promotes the online map---a FoV sliding-window tracking aid---to the final reconstruction, so that refinements in calibration and poses translate into gains in appearance fidelity.
The stage re-initializes the Gaussian model from the refined keyframe schema with all online constraints lifted, and executes global densification, pruning, and spherical-harmonics order elevation.
It can make the 3DGS map more photorealistic and metrically consistent.

\item \textbf{A relocalization module for the robot SLAM lifecycle.}
Our method recovers the global pose from the 3DGS map through an XFeat bypass map, RANSAC-PnP coarse localization, and Gaussian RGB-D warp refinement, complemented by an ICP-based geometric route that registers the query LiDAR scan against the map point cloud and keeps pose recovery unaffected by scene texture.
Based on the globally optimized and reconstructed 3DGS map, the re-embedded pose can be estimated accurately enough to resume mapping and support downstream tasks.

\item \textbf{A calibration-quality-controlled mapping framework.}
Our method makes the geometry that supervises Gaussian training trustworthy by evaluating the quality of the extrinsic calibration offline.
The extrinsic is estimated by an adopted direct-method LiDAR--camera alignment \cite{koide2023calibration} over dedicated calibration segments.
The residuals will be written into every keyframe's depth--color binding and pose filtering will absorb them only up to the observability limit.
Under the calibrated geometry, keyframe poses and Gaussian map parameters will be refined alternately under a gated differentiable-rendering and LiDAR-projection objective.
\end{enumerate}

\subsection{Paper Organization}
\label{sec:intro-org}
The remainder of this paper is organized as follows.
Section~\ref{sec:rw} reviews related works, covering the background of 3D Gaussian splatting and the recent 3DGS SLAM systems.
Section~\ref{sec:method} presents the method: the system overview and the three main modules: the calibration refinement module, the global reconstruction module and the relocalization module.
Section~\ref{sec:exp} reports the experiments, including the setup, the extrinsic calibration evaluation, the online and offline reconstruction evaluation, and the relocalization evaluation.
Section~\ref{sec:concl} concludes the paper and outlines some future directions.

\section{Related Works}
\label{sec:rw}

\subsection{3D Gaussian Splatting}
\label{sec:rw-3dgs}
Neural radiance fields (NeRF) \cite{mildenhall2020nerf} achieved photorealistic novel view synthesis by optimizing a continuous volumetric function through differentiable ray sampling.
However, the dense per-ray sampling makes training and rendering computationally expensive and thus precludes its real-time use.
As an alternative, 3D Gaussian splatting (3DGS) \cite{kerbl3Dgaussians} represents the scene explicitly as anisotropic Gaussian ellipsoids and renders them by differentiable rasterization, attaining real-time rendering at quality competitive with NeRF while keeping the representation explicit and editable.

Image-only 3DGS reconstruction builds on this representation under purely photometric supervision, with camera poses and initial points typically supplied by structure-from-motion \cite{schoenberger2016sfm}.
Subsequent works relaxed these requirements: CF-3DGS \cite{fu2024cf3dgs} progressively grows the Gaussian map frame by frame from an unposed video stream, and InstantSplat \cite{fan2024instantsplat} jointly optimizes camera poses and Gaussian parameters from sparse unposed images.
However, without metric depth, the geometric accuracy of these systems is bounded by photometric evidence alone.
Recently, many research efforts have focused on coupling cameras with LiDAR to bring metric geometry into 3DGS reconstruction.
GS-LIVO \cite{hong2025gslivo} integrates LiDAR--inertial--visual odometry with real-time Gaussian mapping, Gaussian-LIC2 \cite{lang2025gaussianlic2} couples a continuous-time LiDAR--inertial--camera front end with incremental 3DGS mapping, and LiV-GS \cite{xiao2025livgs} aligns LiDAR point clouds directly with differentiable Gaussian maps for large-scale outdoor SLAM.

As an important application, 3DGS has also entered the SLAM loop on the camera-only side for robots.
SplaTAM \cite{keetha2024splatam} formulates RGB-D SLAM with silhouette-guided densification, Gaussian Splatting SLAM \cite{matsuki2024gaussian} performs geometric verification in a monocular Gaussian map, and Photo-SLAM \cite{huang2024photoslam} couples ORB-SLAM3 tracking with primitive-based Gaussian mapping.

\subsection{LiDAR--Inertial--Visual SLAM}
\label{sec:rw-livo}

In real-world robotic applications, SLAM systems that fuse LiDAR, inertial, and visual observations have been widely studied in the literature and are increasingly deployed.
LIV SLAM supplies the metric backbone of such systems---real-time poses, keyframes, and depth.
The front end keeps its accuracy within a real-time budget.
Most research focused on the front-end odometry, with the back end limited to pose graph optimization and loop closure.
FAST-LIO2 \cite{xu2022fastlio2} sets the standard for IESKF LiDAR--inertial odometry with an ikd-Tree map, and FAST-LIVO2 \cite{zheng2025fastlivo2} improved it and achieved tightly coupled LIV fusion with a unified voxel map and error-state iterated Kalman sequential updates.
R3LIVE \cite{lin2022r3live} performs real-time RGB point-cloud mapping by combining LIO geometry with visual-inertial texture.
LVI-SAM \cite{shan2021lvisam} couples visual-inertial and LiDAR-inertial subsystems on a factor graph.
Then, GS-LIVO \cite{hong2025gslivo} is proposed as the first LiDAR--inertial--visual odometry system with real-time Gaussian mapping and is our direct system basis.

In such systems, the LiDAR geometry and camera photometry are registered into one common frame by the LiDAR--camera extrinsic calibration.
So, the calibration must be faithful enough that extrinsic and timing residuals do not corrupt the depth--color bindings that supervise Gaussian training.
Most target-based methods require dedicated calibration boards whose materials and placement must be engineered for simultaneous LiDAR and camera visibility.
On the other hand, targetless LiDAR--camera calibration methods \cite{koide2023calibration,pandey2012mutual,li2023joint} have also been proposed by treating extrinsic estimation as an independent offline preprocessing step, decoupled from mapping.
The direct alignment of dense LiDAR geometry against image appearance over dedicated calibration segments would be a correct path to a better extrinsic.
However, the calibration quality has not yet been analyzed in terms of how it propagates into the mapping and relocalization.
On most public datasets, the dedicated calibration data are also not released, and the calibration chain must be reconstructed from the mapping bags by the users themselves.
In the vision SLAM literature, pose-free Gaussian methods such as CF-3DGS \cite{fu2024cf3dgs} and InstantSplat \cite{fan2024instantsplat} jointly optimize poses and 3DGS from unposed video streams or sparse image sets.
However, they also only rely on photometric energy alone and lack metric supervision.

\subsection{Online and Offline 3DGS Mapping}
\label{sec:rw-online}

In the mentioned LIV systems, the 3D Gaussian map is always built with two mapping methods: one is an online method by building the map with the real-time tracker; and the other one is an offline method by rebuilding the maps after the run.
Online mapping grows the map incrementally alongside the front end.
GS-LIVO \cite{hong2025gslivo} and Gaussian-LIC2 \cite{lang2025gaussianlic2} update the Gaussians together with LiDAR--inertial--visual tracking.
LiV-GS \cite{xiao2025livgs} aligns sparse LiDAR scans directly with differentiable Gaussian maps in large-scale outdoor scenes.
These online maps could be generated under tight budgets, with limited densification, low-order spherical harmonics, and sliding-window optimization.
However, the accuracy and reliability are limited by the real-time constraint.
In order to avoid costly retraining, LIVE-GS \cite{park2025livegs} maintains global consistency of the online map through surfel-level loop closure, and the Gaussians affected by the corrected poses are updated by rigid transformation.
However, its pose refinement relies on geometric registration only, without rendering feedback.
On the other hand, offline refinement rebuilds the map after the run, without considering the real-time constraint, and the poses can be refined jointly with the Gaussian parameters.
Peng et al. proposed a method to refine 3DGS from coarsely posed images and noisy LiDAR point clouds by constrained optimization, jointly optimizing camera intrinsics, poses, and Gaussian parameters \cite{peng2025constrained}.
However, the calibration quality is still not treated as a variable of reconstruction quality.
A refinement stage is required to lift the online constraints, rebuild the keyframe schema into the final reconstruction, and propagate the benefit of calibration explicitly through global reconstruction.

\subsection{Relocalization}
\label{sec:rw-reloc}

Re-enterability is likewise essential for SLAM: tracking loss is inevitable in real deployments, caused by fast rotation, motion blur, illumination changes, or dynamic occlusion. 
Without pose recovery, the accumulated map becomes orphaned precisely when downstream tasks depend on it.
Classical robot relocalization recovers the pose by retrieving keyframes with bag-of-words descriptors and solving PnP, as in the relocalization module used in ORB-SLAM3 \cite{campos2021orbslam3}.
Some learning-based methods remove the explicit feature pipeline: ACE \cite{brachmann2023ace} regresses scene coordinates and trains its map within minutes.
On the 3DGS side, GSplatLoc \cite{sidorov2024gsplatloc} grounds XFeat descriptors into a 3DGS map for visual localization, combining 2D--3D coarse localization with photometric warp refinement, and is the direct reference for our module.
However, these works address localization as an isolated pose-estimation problem on benchmark datasets, and none of them builds relocalization upon a calibration-refined 3DGS map or quantifies how extrinsic quality propagates into relocalization reliability.
A relocalization module must be able to recover the pose from the generated 3DGS map, and its reliability must be tied to the calibration quality of the map it queries.

\section{Method}
\label{sec:method}
\subsection{Overview}
\label{sec:overview}

\begin{figure}[t]
\centering
\includegraphics[width=0.95\textwidth]{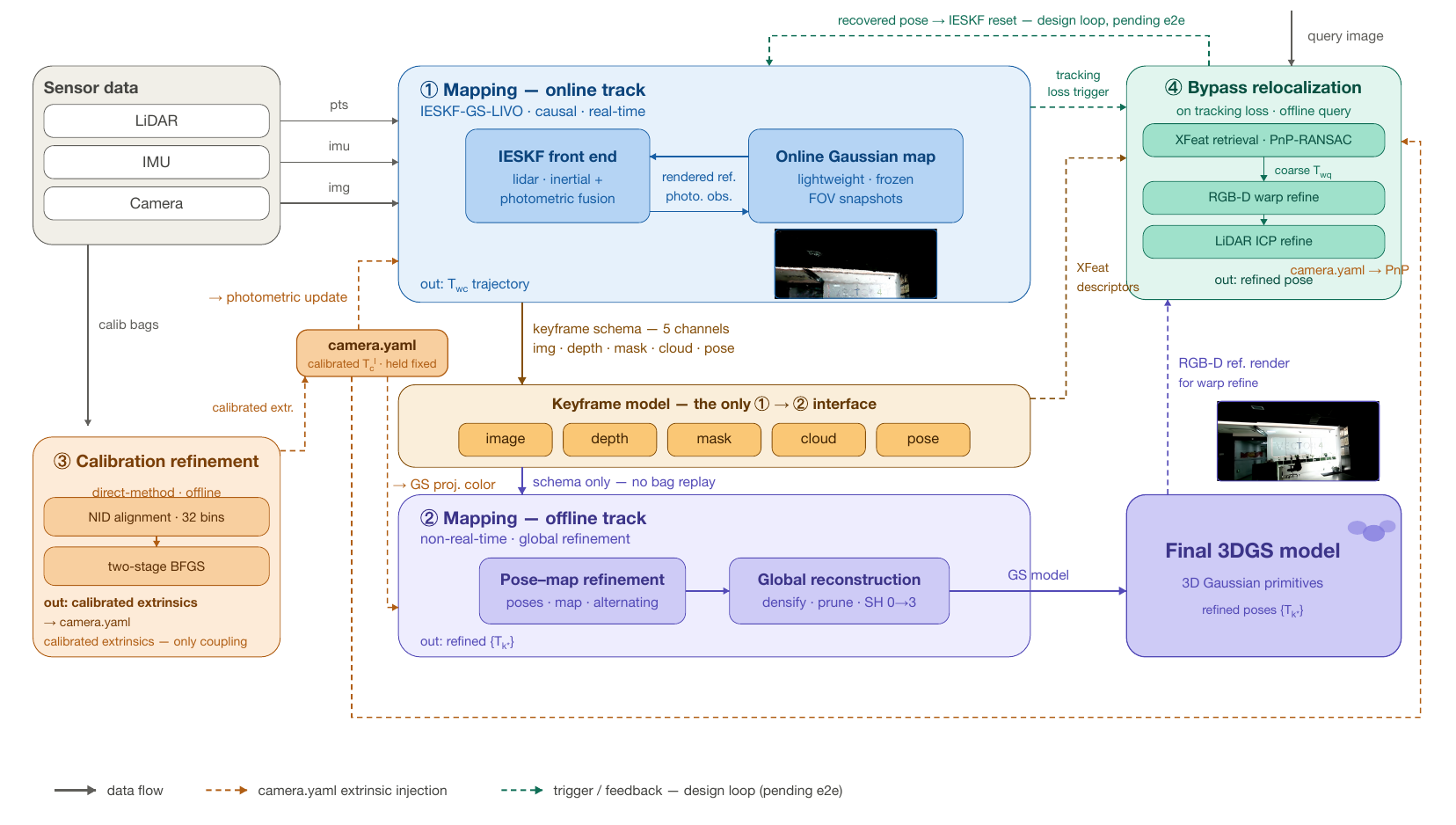}
\caption{\textbf{System overview.}
The online track \protect\textcircled{1} runs an IESKF over LiDAR, inertial, and visual measurements against a compact Gaussian tracking reference, and keyframes accumulate under the five-channel contract into the shared keyframe schema. 
The offline track \protect\textcircled{2} consumes the keyframe schema: joint optimization refines keyframe poses and map, after which global reconstruction re-initializes the Gaussians into the final 3DGS model.
The calibration refinement \protect\textcircled{3} is an independent process: the extrinsic is estimated once per sensor rig by direct-method calibration and reused by both tracks.
The bypass relocalization path \protect\textcircled{4} matches descriptors against the relocalization database upon tracking loss, registers the query LiDAR scan to the map point cloud by ICP, and hands the refined pose back to the front end. 
}
\label{fig:system_overview}
\end{figure}

Figure~\ref{fig:system_overview} illustrates the configuration of the system (GEMOR-GS) that we developed, which is organized as two tracks (online and offline tracks) connected by a keyframe schema.
The sensor data are captured by a rig carrying three rigidly mounted sensor types: a 3D LiDAR producing metric point clouds, a color camera producing images, and a built-in IMU producing high-rate inertial measurements.
The LiDAR--camera extrinsic $\mathbf{T}_c^l$ is estimated offline by a direct-method calibration method \cite{koide2023calibration} and held fixed throughout.
The LiDAR--camera time offset $\delta t$ is also fixed by a hand-measured value.
The online modeling module runs causally in real time from the measured sensor data, and its role is limited to estimating poses and building the keyframe schema.
The keyframe schema is the only interface between the two tracks, with atomic writes and versioning for consistency.
The offline track consumes the keyframe schema without real-time constraints.
The calibration refinement module first refines keyframe poses and the map under the extrinsic estimated by direct-method calibration.
The global reconstruction module then rebuilds the final 3DGS model from the refined keyframe schema.
In practice, a direct-method calibration method is used to estimate the extrinsic once per sensor rig over dedicated calibration segments and reuses it across both online and offline runs.
A bypass relocalization module stands outside the online pipeline, which would be triggered while tracking loss and recover the global pose from the relocalization database, with an ICP-based geometric route that registers the query LiDAR scan to the map point cloud.
Table~\ref{tab:notation} defines the notation used in this paper.

\begin{table}[t]
\centering
\caption{\textbf{Notation.}}
\begin{tabular}{@{}l p{7.6cm}@{}}
\toprule
Symbol & Meaning \\
\midrule
$\mathbf{T}_w^b$ & world-to-body transform (IESKF state) \\
$\mathbf{T}_c^l$ & LiDAR--camera extrinsic, estimated offline by direct-method calibration and held fixed \\
$\delta t$ & LiDAR--camera time offset, a fixed hand-measured value (28~ms) \\
$\mathbf{G} = \{\mu_i, \Sigma_i, \alpha_i, \mathbf{c}_i\}$ & Gaussian map: positions, covariances, opacities, colors \\
$\mathcal{A} = \{(\mathbf{I}_k, \mathbf{D}_k, \mathbf{M}_k, \mathbf{P}_k, \mathbf{T}_k)\}$ & keyframe schema: image, sparse depth, dynamic mask, cloud, pose \\
$\mathcal{R}(\cdot)$ & differentiable rasterization \\
$\pi(\cdot)$ & LiDAR-to-image projection, which depends on $\mathbf{T}_c^l$ \\
\bottomrule
\end{tabular}
\label{tab:notation}
\end{table}

\subsection{Calibration Refinement Module}
\label{sec:calib}
\textbf{Direct-method extrinsic calibration.}
Extrinsic calibration is the baseline of the system.
It will affect all following modules, and its quality will be critical to the overall performance.
Therefore, we first validate the direct-method calibration module and then quantify how extrinsic quality propagates into reconstruction.
In this module, the extrinsic $\mathbf{T}_c^l$ is estimated by a direct-method LiDAR--camera alignment \cite{koide2023calibration}, which is a targetless and direct formulation.
It does not need a calibration board, exploits the ambient scene structure and texture directly, and its pixel-level NID alignment is reported to be more robust than edge-based indirect registration.
The LiDAR point cloud and the image are registered by minimizing the normalized information distance (NID) between LiDAR intensity projections and image gradients.
The estimation is obtained by a two-stage BFGS search started from the rig geometry.
The estimated extrinsic is promoted once and held fixed for following modules: the online track (via a fixed extrinsic input to the filter) and the offline track (via the same value in the keyframe schema).
No extrinsic update will happen at run time.

\textbf{Offline pose--map alternating refinement.}
The base objective refines keyframe poses and the map alternately over the keyframe schema $\mathcal{A}$ as follows:
\begin{equation}
 \min_{\{\mathbf{T}_k\}, \mathbf{G}} \sum_{k}
 \mathcal{L}_{render}\big(\mathcal{R}(\mathbf{G}, \mathbf{T}_k),
 \mathbf{I}_k\big) + \lambda_{reg}\,\mathcal{L}_{reg}(\mathbf{G}),
 \label{eq:base}
\end{equation}
where $\mathcal{L}_{render}$ is the photometric rendering loss and $\mathcal{L}_{reg}$ regularizes the map.
Poses and map are optimized alternately: gradients through the rasterizer refine the poses with the map fixed, and standard Gaussian training refines the map with the poses fixed.
Optimization proceeds in three alternating stages: (1) initial 3DGS training with the poses fixed; (2) refinement of the keyframe poses under the rendering and cross-frame cloud-consistency terms, with the map fixed; (3) retraining of 3DGS under the updated poses.
It will be iterated until convergence.
Here, the nearest-pairing offsets of up to the camera frame interval are handled by a fixed hand-measured time offset with pose interpolation.
A pose update is accepted only when the following three gates are satisfied: (i) the rendered image contains at least $N_{min}$ valid pixels, (ii) the opacity coverage in the current FoV exceeds $\eta_{min}$, and (iii) the update step does not exceed $(\Delta t_{max}, \Delta\theta_{max})$.
Frames failing any above gate will be archived without refinement and will never be allowed to pollute the estimate.
Finally, the refined poses $\{\mathbf{T}_k^*\}$ and the extrinsic $\mathbf{T}_c^{l*}$ are archived into the keyframe schema for other modules.


\subsection{Online Modeling Module}
\label{sec:online}
In this module, the IESKF front end estimates poses in real time and archives keyframes into the shared schema.
The IESKF state estimation inherits the FAST-LIO2-style error-state formulation, in which the error state $\delta\mathbf{x} = [\delta\mathbf{p}, \delta\mathbf{v}, \delta\mathbf{q}, \delta\mathbf{b}_a, \delta\mathbf{b}_g]^T$ is propagated by the IMU and updated by point-to-plane residuals against an ikd-Tree map and, following GS-LIVO \cite{hong2025gslivo}, additionally by photometric residuals rendered from the online Gaussian map.

The online Gaussian map is initialized by direct insertion of LiDAR returns, one Gaussian per return, and colored from the paired image, without densification.
Each frame, only the FoV-visible subset is micro-optimized under a fixed budget of Adam steps $N_{max}$ (e.g. 50 steps).
After each frame's training, we freeze the current-FoV Gaussian snapshot $\mathbf{G}_t^{frozen}$.
Subsequent frames render only against frozen snapshots, and current-frame Gaussians will never retroactively supervise already-observed frames.

Then, the keyframes will be selected by pose-change thresholds on translation and rotation with a time-interval floor.
Each archived keyframe will also be stored with five channels: the rectified image $\mathbf{I}_k$, the LiDAR-projected sparse depth $\mathbf{D}_k$, the dynamic mask $\mathbf{M}_k$, the rectified world-frame cloud $\mathbf{P}_k$, and the pose $\mathbf{T}_k$.
The latter will be the quantity refined in the following offline stage.

\subsection{Global Reconstruction Module}
\label{sec:global}
In this module, the refined keyframe schema is consumed to rebuild the online Gaussian model into the final reconstruction.
We re-initialize the Gaussian model from the refined keyframe schema (poses $\{\mathbf{T}_k^*\}$, extrinsic $\mathbf{T}_c^{l*}$, and clouds $\{\mathbf{P}_k\}$), where the poses come from the pose--map refinement.
The extrinsic comes from the direct-method calibration module and is also held fixed.
The sparse depth maps will be re-projected under the calibrated extrinsic, so the calibrated geometry directly determines the depth supervision.

During the reconstruction, we lift all online constraints.
Global adaptive densification will proceed without considering the real-time budget.
Pruning will be executed globally on the opacity criterion.
Spherical harmonics will be elevated in stages from degree 0 through degree 3 to represent view-dependent appearance.

The retraining objective uses the same rendering loss defined by \eqref{eq:base}.
Each training view contributes the photometric term $\mathcal{L}_{render}$ which is computed with the same mean-absolute-error ($\mathcal{L}_1$) distance and structural-similarity kernel as the evaluation.
A re-projection depth term that supervises the rendered depth against the sparse depth $\mathbf{D}_k$ is added with a confidence-weighted robust kernel.
The depth weight follows a decay schedule.
The depth term acts as an early geometric prior and hands over to the photometric term as training converges.
The only optimization variable in this module is the Gaussian map $\mathbf{G}$: the keyframe poses $\{\mathbf{T}_k^*\}$ are held fixed as refined by the pose--map refinement of Section~\ref{sec:calib}, and the extrinsic $\mathbf{T}_c^{l*}$ is provided by the direct-method calibration and is not a variable of this module.
Therefore, the calibrated extrinsic directly determines the depth supervision of this module.


\subsection{Relocalization Module}
\label{sec:reloc}
In this module, the pose can be recovered from the generated 3DGS map from a query image if the robot loses tracking.

Based on the final reconstructed 3DGS map, a lightweight relocalization database will be built once from the archived keyframes.
The relocalization only queries the database without touching the 3DGS training.
From the query image, the XFeat \cite{potje2024xfeat} keypoints will be back-projected through the archived depth, and masked by the dynamic mask $\mathbf{M}_k$.
The back-projected point is anchored to its nearest Gaussian.
The high-dimensional descriptors are fused over observations into the relocalization database.
Landmarks below a minimum observation count are discarded, and the database keeps only repeatedly confirmed geometry.

By matching a query image's descriptors against the database and feeding the 2D--3D correspondences to a RANSAC-PnP solver, we first yield a coarse pose $\mathbf{T}_{coarse}$ sufficient to seed the filter reset.
From this $\mathbf{T}_{coarse}$, we then render an RGB-D frame from the 3DGS map and refine the pose by direct RGB-D warp alignment, yielding $\mathbf{T}_{fine}$ at strong-observation quality, which would be output as the pose of the input query image.
Then, the geometric route registers the query LiDAR scan against the map point cloud by iterative closest point (ICP) \cite{besl1992icp}, recovering the global pose from map geometry alone and reducing the effect of scene texture.

\section{Experiments}
\label{sec:exp}
In the experiment, we evaluate three objectives (geometric accuracy, appearance fidelity, and re-enterability), by evaluating the performance of three modules in our method (calibration refinement, global reconstruction, and relocalization).
For the global reconstruction experiment, we also quantify the systematic price of the online budget to verify the necessity of the offline stage by comparing it with the online map.

\subsection{Experimental Setup}
\label{sec:setup}

\begin{figure}[t]
\centering
\includegraphics[width=0.9\textwidth]{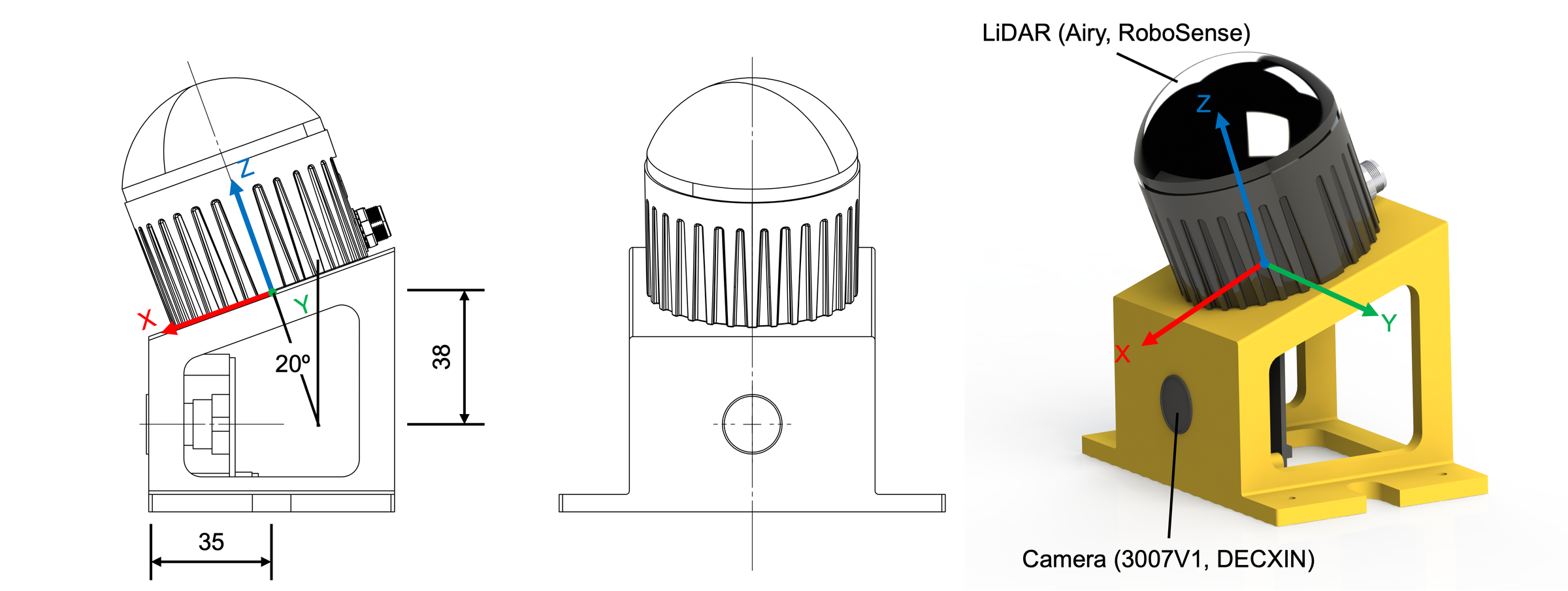} 
\caption{\textbf{Sensor configuration for the self-collected sequence.}
The rig carries a RoboSense Airy 96 LiDAR, a DECXIN-3007V1 color camera, and the Airy built-in IMU.
The LiDAR is mounted on the top of the camera with a 28~mm offset and 20$^\circ$ downward tilt, and the camera is mounted with 25~mm offset front of the LiDAR.
}
\label{fig:rig}
\end{figure}

\begin{table}[H]
\centering
\caption{\textbf{Sequences used in the experiments.}}
\footnotesize
\begin{tabular}{@{}l l l l l l c@{}}
\toprule
Sequence & Dataset & Scene & LiDAR & Camera & IMU & Keyframes \\
\midrule
Ours & self-collected & indoor & RoboSense Airy 96 & DECXIN-3007V1 & Built-in & 837 \\
CBD & FAST-LIVO2 \cite{zheng2025fastlivo2} & outdoor & Livox Avia & MV-CA013-21UC & Built-in & 1117 \\
Retail & FAST-LIVO2 \cite{zheng2025fastlivo2} & outdoor & Livox Avia & MV-CA013-21UC & Built-in & 1286 \\
Bright & FAST-LIVO2 \cite{zheng2025fastlivo2} & indoor & Livox Avia & MV-CA013-21UC & Built-in & 596 \\
hku\_campus & R3LIVE \cite{lin2022r3live} & indoor & Livox Avia & MV-CA013-21UC & Built-in & 1958 \\
tuhh\_day\_04 & MCD \cite{nguyen2024mcd} & outdoor & Ouster OS1-64 & RealSense D455b & VN200 & 838 \\
\bottomrule
\end{tabular}
\label{tab:datasets}
\end{table}

\begin{table}[H]
\centering
\caption{\textbf{Sensor specifications.}}
\footnotesize
\begin{tabular}{@{}l l p{4.9cm} l l@{}}
\toprule
Sensor & Maker & Key specs & Resolution & Rate \\
\midrule
RoboSense Airy 96 & RoboSense & mechanical, 360$^\circ\times$90$^\circ$ FoV, 1.72M pts/s, 1~cm at $1\sigma$ & -- & 10~Hz \\
Livox Avia & Livox & solid-state, non-repetitive scan, 70.4$^\circ\times$77.2$^\circ$ FoV & -- & 10~Hz \\
Ouster OS1-64 & Ouster & mechanical, 64 lines & 64$\times$1024 & 10~Hz \\
\midrule
DECXIN-3007V1 & DECXIN & color, rolling shutter & 1280$\times$720 & 30~Hz \\
MV-CA013-21UC & HIKROBOT & color, global shutter & 1280$\times$1024 & 10/15~Hz \\
RealSense D455b & Intel & color stereo depth camera & 640$\times$480 & 30~Hz \\
\midrule
Airy built-in & RoboSense & shares the LiDAR time base & -- & -- \\
BMI088 (Avia built-in) & Bosch & -- & -- & 200~Hz \\
VectorNav VN200 & VectorNav & external & -- & 400~Hz \\
\bottomrule
\end{tabular}
\label{tab:sensors}
\end{table}

We evaluated our method on six sequences from one self-collected and five public indoor and outdoor scenes to show the generalizability of our method even for different combinations of sensors.
Table~\ref{tab:datasets} lists the sequences and their sensor suites.
These sequences covered a wide range of indoor and outdoor scenes, with different lighting conditions, dynamic objects, and occlusions.
Figure~\ref{fig:rig} shows the configuration of the sensor rig that we designed for collecting our sequence.
The collected sequence is a 90~s indoor exhibition-hall scene, recorded with a hemispherical LiDAR (Airy, RoboSense) whose built-in IMU shares the LiDAR time base and a color camera (3007V1, DECXIN) with fixed frame rate (30~Hz).
Among the sequences, Bright (an indoor screen wall) and hku\_campus (an indoor university corridor) are indoor scenes, while CBD (a building block with high-rise facades), Retail (a street with close-range pedestrians and occluders), and tuhh\_day\_04 (a large outdoor campus) are outdoor scenes.

For our sensor rig, the extrinsic is estimated by the direct-method calibration of Section~\ref{sec:calib} over 13 dedicated calibration segments and held fixed thereafter.
The other five public sequences reuse the extrinsics published with their datasets for reconstruction and relocalization; their dedicated calibration data are not released, so the calibration module is validated on segments cut from the same bags, with the released extrinsics as the initialization reference (Section~\ref{sec:exp-joint}).
We also applied some injection to other five public sequences to test the effectiveness of the calibration refinement and online--offline modelling modules to confirm the validity.

For evaluating the result, in following experiments, the rendering quality is measured by the peak signal-to-noise ratio ($\mathrm{PSNR}$) and the structural similarity index ($\mathrm{SSIM}$) \cite{wang2004ssim}, computed as follows:
\begin{align}
\mathrm{PSNR} &= 10 \log_{10} \frac{I_{\max}^{2}}{\mathrm{MSE}}, \\
\mathrm{SSIM}(\mathbf{x}, \mathbf{y}) &= \frac{(2\mu_x\mu_y + c_1)(2\sigma_{xy} + c_2)}{(\mu_x^2 + \mu_y^2 + c_1)(\sigma_x^2 + \sigma_y^2 + c_2)}.
\end{align}
where $\mathrm{MSE}$ is the mean squared error over rendered and ground-truth frames; $I_{\max}$ is the maximum pixel value; $\mu, \sigma, c_1, c_2$ are the local means, variances, covariance, and $\mathrm{SSIM}$ is the stabilizing constants.
$\mathrm{PSNR}$ uses the same CUDA kernel as offline training and $\mathrm{SSIM}$ the FusedSsim implementation.
Calibration accuracy is measured by the rotation and translation errors in recovery relative to the reference calibration.
Relocalization is evaluated by the success rate, the PnP inlier ratio, and the reprojection RMSE in pixels.
The geometric ICP route is additionally evaluated by its success count and the median translation (cm) and rotation ($^\circ$) errors of the registered poses against the reference trajectories.
Map quality is quantified by the PSNR of rendered images, which reflects what relocalization inherits from the extrinsic calibration.

We implemented and executed all experiments on a server computer with 8 NVIDIA GeForce RTX 4090 GPUs (24~GB) under CUDA 12.8.
The online micro-optimization uses its fixed budget of 50 Adam steps per frame (Section~\ref{sec:online}), the XFeat relocalization engine uses its native 64-dimensional descriptors (Section~\ref{sec:reloc}), and the offline reconstruction uses a 100k final-reconstruction budget (Section~\ref{sec:global}).

\subsection{Experiment 1: Extrinsic Calibration Evaluation}
\label{sec:exp-joint}

\textbf{Calibration module validation.}
For our self-collected sequence, there is no independent hardware ground truth for the extrinsic.
We estimated it using the calibration module and validated by an injection--recovery protocol .
The rig construction data serve as the initial value of the calibration.
A known offset is injected into this initial value, and the recovery error of the output is measured against the calibration reference, which is the extrinsic actually consumed by every run of this system.
For other five public sequences, same protocol is applied.
We use their released dataset extrinsics as the initial value, and inject the same offset to test the calibration module.

The NID is computed over a 32-bin discretization in all calibration runs.
For our self-collected sequence, 13 dedicated calibration segments of 12~s each are cut from the bag.
On the public sequences, 3--4 calibration segments of 12~s each are cut from the same bags
The estimator runs with the identical configuration simultaneously on all six sequences.
The released extrinsics are not guaranteed to be the NID optimum.
They embed their own protocol differences and, on Retail, an unpublished camera--LiDAR time offset.
Therefore, the primary measure is unified as the consistency between the output of the injected run and that of the uninjected reference run.

\begin{table}[t]
\centering
\caption{\textbf{Result of extrinsic calibration evaluation.}
Use the direct-method calibration recovery over injected initial offsets on all six sequences.
}
\small
\setlength{\tabcolsep}{4pt}
\begin{tabular}{@{}l c c c c@{}}
\toprule
Sequence & Segments & $\theta$ ($+1.51^\circ$) & $\|\Delta t\|$ & $\theta$ ($+5.0^\circ$) \\
\midrule
Ours & 13 & $0.081^\circ$ (94.6\%) & 4.3~mm & $0.099^\circ$ (98.0\%) \\
tuhh\_day\_04 & 4 & $0.025^\circ$ & 6.1~mm & $0.889^\circ$ (82.2\%) \\
CBD & 3 & $0.078^\circ$ & 25.8~mm & $0.073^\circ$ (98.5\%) \\
hku\_campus & 4 & $0.059^\circ$ & 10.4~mm & $2.417^\circ$ (51.7\%)$^{\uparrow}$ \\
Retail & 3 & $0.227^\circ$ & 39.9~mm & $1.335^\circ$$^{\uparrow}$ \\
Bright & 3 & $4.626^\circ$$^{\uparrow}$ & 865.2~mm$^{\uparrow}$ & $8.040^\circ$$^{\uparrow}$ \\
\bottomrule
\end{tabular}
\label{tab:recover}
\end{table}

Table~\ref{tab:recover} shows the recovery on all six sequences.
The rotation $\theta$ and translation $\|\Delta t\|$ are the discrepancies from the reference run.
The rotation discrepancies from the uninjected outputs, and the parenthesized rates are the recovery percentages relative to the injected dose ($= 1 - \theta / \text{dose}$).
$^{\uparrow}$ marks degraded or failed recovery, where the output is too far away from the reference run by more than $0.15^\circ$.
Four of the six sequences (ours, tuhh\_day\_04, CBD, and hku\_campus) recover the $+1.51^\circ$ dose fully.
Ours recovers to $0.081^\circ$ (94.6\%) with a 4.3~mm translation residual, and the three public sequences ---tuhh\_day\_04, CBD, and hku\_campus---converge to the same basin as their reference runs at $0.025$--$0.078^\circ$ and 6.1--25.8~mm, with NID-cost differences below 0.001.
Retail converges self-consistently ($0.227^\circ$ between the two outputs), but both solutions lie about $2^\circ$ from the released value, a reference gap most plausibly attributable to the unpublished time offset; the recovery mechanism itself operates normally.
Our method is failure on the sequence Bright, where the $+5.0^\circ$ injection fails to recover to $4.626^\circ$ (865.2~mm), due to the extreme weak-geometry scene with only three segments.

The remaining injection groups replicate these findings across doses and offsets.
On self-collected segments, the $+5.0^\circ$ injection recovers it to $0.099^\circ$ (98.0\%) on 13 segments.
On the public segments it holds on CBD ($0.073^\circ$, 98.5\%) and tuhh\_day\_04 ($0.889^\circ$, 82.2\%), is partial on hku\_campus ($2.417^\circ$, 51.7\%) and Retail ($1.335^\circ$), and fails on Bright.
The reliable-recovery dose boundary lies below the $5^\circ$ basin of the 13 self-collected segments.
The $+0.153^\circ$ output marks the absolute accuracy floor of the estimator: $0.064^\circ$ on Ours, and $0.015$--$0.066^\circ$ on the recovered public sequences (CBD, tuhh\_day\_04, hku\_campus).
A $+50$~mm translation injection alone is recovered to 11.4~mm (77\%) on Ours, consistent with the weaker NID constraint along the optical axis; on the recovered public sequences the translation output returns to the reference-run neighborhood (2.7--39.5~mm) and rotation is perturbed by at most $0.15^\circ$.
Under the combined rotation-and-translation injection, recovery stays at the rotation-only level---95.1\% on Ours and 95.9--97.0\% on the recovered public sequences.
Therefore, we conclude that the rotation and translation gradients are decoupled rather than competing.
The failure pattern of Bright and the reference gap of Retail repeat across all groups.
We consider the failure as a result of the multi-basin cost surface or the unpublished time offset, not a failure of the estimator itself.
These injection--recovery results show that the estimator is effective and robust for extrinsic calibration, and suitable for major sequences with sufficient geometry and texture.

\begin{table}[t]
\centering
\caption{\textbf{Extrinsic dose response.}
The $\mathrm{PSNR}$ of the final reconstruction under the control, low-dose ($+0.153^\circ$), and high-dose ($+1.51^\circ$) extrinsic rotation.
}
\small
\begin{tabular}{@{}l c c c@{}}
\toprule
\multirow{2}{*}{Sequence} & \multicolumn{3}{c}{$\mathrm{PSNR}$ (dB)} \\
\cmidrule(l){2-4}
 & Control & Low dose ($+0.153^\circ$) & High dose ($+1.51^\circ$) \\
\midrule
Ours & 30.06 & 30.34 ($+0.28$) & 29.22 ($-0.84$) \\
Retail & 33.92 & 33.73 ($-0.19$) & 33.17 ($-0.76$) \\
CBD & 33.76 & 33.63 ($-0.14$) & 31.74 ($-2.02$) \\
tuhh\_day\_04 & 27.96 & 27.54 ($-0.41$) & 26.79 ($-1.17$) \\
hku\_campus & 29.80 & 29.66 ($-0.14$) & 28.93 ($-0.86$) \\
Bright & 32.91 & 33.63 ($+0.72$) & 34.75 ($+1.84$) \\
\bottomrule
\end{tabular}
\label{tab:dose}
\end{table}

We also evaluate the effect of the extrinsic on the reconstruction and relocalization quality by injecting a $+0.153^\circ$ and $+1.51^\circ$ rotation about the camera-frame $y$ axis into the extrinsic.
At the same time, all other parameters (e.g., camera intrinsics, $\delta t$, IESKF parameters, and iteration budget) are held identical.
In total, 18 runs are executed: a control, a low-dose, and a high-dose configuration for six sequences each.
Table~\ref{tab:dose} reports the dose response over 18 runs.
Averaged over the six sequences, the low dose leaves $\mathrm{PSNR}$ at the control level (31.42 versus 31.40~dB), whereas the high dose degrades the mean by 0.6~dB (30.77 versus 31.40~dB).
For each sequence, the low dose improves Ours ($+0.28$~dB) and Bright ($+0.72$~dB) and slightly degrades for all others (Retail ($-0.19$~dB), CBD ($-0.14$~dB), hku\_campus ($-0.14$~dB), and tuhh\_day\_04 ($-0.41$~dB)).
The high dose degrades five sequences, CBD ($-2.02$~dB), tuhh\_day\_04 ($-1.17$~dB), hku\_campus ($-0.86$~dB), Ours ($-0.84$~dB), and Retail ($-0.76$~dB), and only improves for Bright ($+1.84$~dB).
$\mathrm{SSIM}$ also corroborates the direction of these changes: the high-dose $\mathrm{SSIM}$ falls below control on all normal sequences, with 0.933 versus 0.949 on CBD, 0.874 versus 0.900 on tuhh\_day\_04, and 0.886 versus 0.898 on hku\_campus.
Across the six sequences, the high dose degrades $\mathrm{PSNR}$ by 0.76 to 2.02~dB in five of six sequences, while the low dose stays within the noise band of $-0.41$ to $+0.28$~dB, placing the detectable threshold near 1.5$^\circ$.
The single exception is also for sequence Bright (the only extreme weak-texture sequence), where a flat loss surface lets the pose and Gaussian-shape degrees of freedom absorb the extrinsic bias, as corroborated by its 0\% relocalization success.
The dose-response results establish that the extrinsic is a sensitive factor in reconstruction quality, with damage onset near 1.5$^\circ$.
The calibration delivered by the direct method that we used in this work is suitable for refining the extrinsic of most sequences in other modules.

\subsection{Experiment 2: Online and Offline Reconstruction Evaluation}
\label{sec:exp-recon}
In this experiment, using the extrinsic data from the above calibration method, we generate the final global reconstruction.
The quality was evaluated by comparing it with the online map.
As the upstream baseline, it was also compared with the map produced by Gaussian-LIC2 \cite{lang2025gaussianlic2}, which is a wild-known method that has been confirmed in many other research works.
In order to compare the online map and the offline reconstruction under identical inputs, we replay the online mapping algorithm offline with same replay tool.
Every input to the online mapper---the rectified image, the world-frame cloud, and the pose---is byte-identical to what the online run.
The hyperparameters are also identical for the all run configurations
Note that, the replay is a faithful offline re-enactment of the online mapping, so the comparison isolates the algorithmic quality of the map rather than real-time overhead.

\begin{table}[t]
\centering
\caption{\textbf{Quality of the offline global reconstruction.}}
\begin{tabular}{lccc}
\toprule
Sequence & Keyframes & Offline $\mathrm{PSNR}$ (dB) & Offline $\mathrm{SSIM}$ \\
\midrule
Ours & 837 & 30.31 & 0.946 \\
CBD & 1117 & 33.83 & 0.949 \\
Retail & 1286 & 34.07 & 0.957 \\
Bright & 596 & 32.89 & 0.944 \\
hku\_campus & 1958 & 30.31 & 0.904 \\
tuhh\_day\_04 & 838 & 28.09 & 0.901 \\
\bottomrule
\end{tabular}
\label{tab:global}
\end{table}

\begin{table}[t]
\centering
\caption{\textbf{Comparison of online map and offline global reconstruction.}}
\begin{tabular}{lccccc}
\toprule
Sequence & Pairs & Online $\mathrm{PSNR}$ (dB) & Offline $\mathrm{PSNR}$ (dB) & $\Delta$ mean (dB) & Win rate \\
\midrule
Ours & 837 & 15.92 & 30.31 & $-14.39$ & 100\% \\
CBD & 1117 & 13.90 & 33.83 & $-19.93$ & 100\% \\
Retail & 1286 & 14.57 & 34.07 & $-19.50$ & 100\% \\
Bright & 596 & 17.93 & 32.89 & $-14.96$ & 100\% \\
hku\_campus & 1958 & 13.19 & 30.31 & $-17.11$ & 100\% \\
tuhh\_day\_04 & 838 & 13.83 & 28.09 & $-14.26$ & 100\% \\
\bottomrule
\end{tabular}
\label{tab:online}
\end{table}

\begin{figure}[t]
\centering
\includegraphics[width=0.95\textwidth]{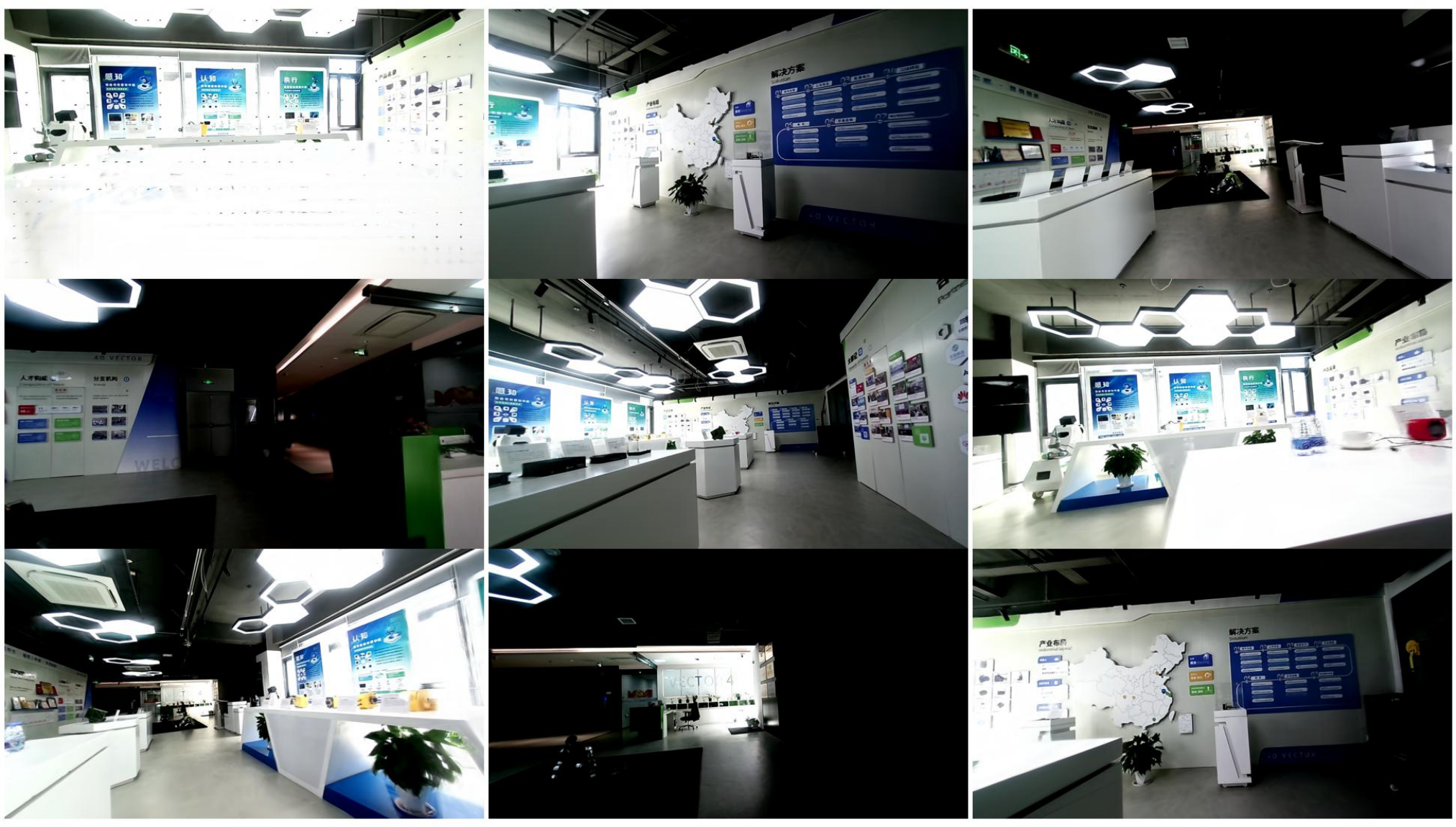}
\caption{\textbf{Examples of offline global reconstruction on the self-collected sequence.}
Rendered training views from the final global 3DGS model after calibration-refined reconstruction.}
\label{fig:teaser}
\end{figure}

Table~\ref{tab:global} shows the offline final states on the six sequences, and Table~\ref{tab:online} shows the paired online--offline comparison.
All compared keyframes of each run are paired by index and no sampling.
The offline reconstruction outperforms its online counterpart in all 6,632 pairs, with a per-frame PSNR gap of 14.3--19.9~dB. 
Figure~\ref{fig:teaser} shows the offline reconstruction that our system produces on the self-collected exhibition-hall sequence.
The two training budgets differ by a factor of 3--6 (42k steps for online modeling versus 140--240k for offline reconstruction), both with full densification and pruning; the offline reconstruction grows from 20k to as many as 2.36M Gaussians.

To attribute this gap, we evaluate the refined online map at two time points.
First, each frame is rendered once when first observed mid-run.
Second, each frame is rendered once after the map is frozen. 
Both measurements thus come from the same online model rather than from two separately trained models.

We further decompose the $\mathrm{PSNR}$ gap into two separable contributions. 
The first is FoV sliding-window forgetting: mid-run evaluation on freshly observed frames reaches 21.8--27.2~dB, whereas final review with the frozen map yields 13.2--17.9~dB, a drop of 6--14~dB per sequence.
The second is the absence of global reconstruction, which accounts for the remaining gap from final review to the offline state.


\begin{table}[t]
\centering
\caption{\textbf{Training-view calibrated $\mathrm{PSNR}$ against Gaussian-LIC2 on
the same bags.}}
\small
\begin{tabular}{@{}p{2.6cm} p{2.4cm} p{2.2cm} c p{2.9cm}@{}}
\toprule
Sequence & Ours, control (dB) & LIC2 (dB) & $\Delta$ (dB) & LIC2 trajectory drift \\
\midrule
CBD & 33.65 at 1280$\times$1024 & 25.62 at 640$\times$512, JPEG & $+8.03$ & 0.029~m / 7.10$^\circ$ \\
Retail & 34.05 at 640$\times$512 & 24.62 at 640$\times$512 & $+9.43$ & 0.035~m / 0.55$^\circ$ \\
Bright & 32.59 at 1280$\times$1024 & 25.51 at 640$\times$512 & $+7.08$ & 0.321~m / 2.80$^\circ$ \\
hku\_campus (R3LIVE) & 30.21 at 1280$\times$1024 & 24.83 at 640$\times$512 & $+5.38$ & 0.048~m / 4.23$^\circ$ \\
tuhh\_day\_04 (MCD, 90~s) & 28.00 at 640$\times$480 & 19.22 at 640$\times$480 & $+8.78$ & 121~m / 124$^\circ$, tracking lost \\
Ours & 30.27 at 1280$\times$720 & 17.98 at 1280$\times$720 & $+12.29$ & 1.27~m / 77$^\circ$, drift \\
\bottomrule
\end{tabular}
\label{tab:lic2}
\end{table}

\begin{figure}[p]
\centering
\includegraphics[width=0.95\textwidth]{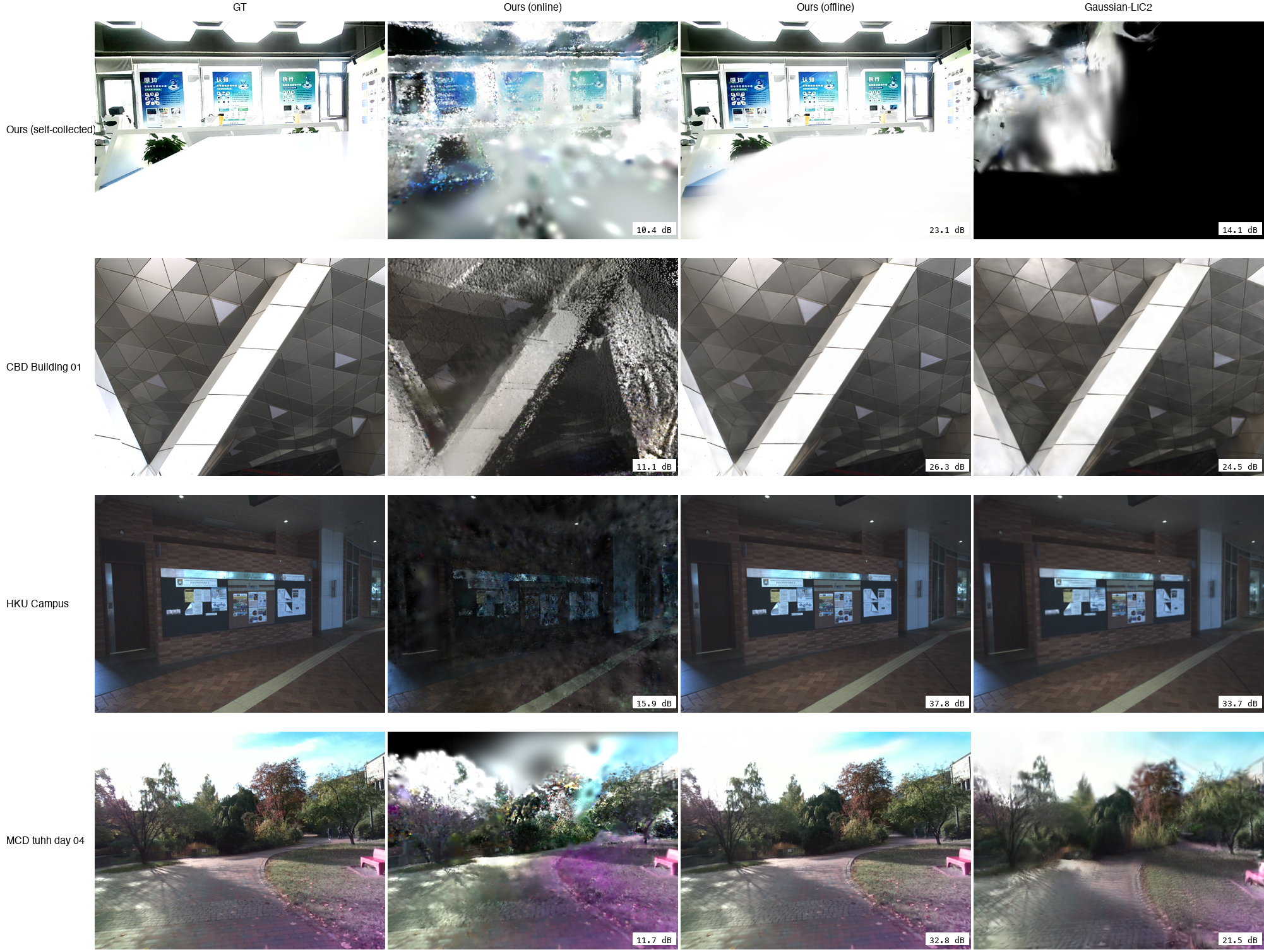}
\caption{\textbf{Qualitative comparison against Gaussian-LIC2 on four
sequences.} One aligned keyframe per sequence (rows), rendered from
matched poses: ground-truth camera images, our online Gaussian map
(tracking reference, rendered at the end of the online stage), our
calibration-refined global reconstruction, and the upstream
Gaussian-LIC2 map. Per-cell $\mathrm{PSNR}$ is computed against the corresponding
ground-truth image at native resolution (ours against our keyframe GT,
LIC2 against its own same-pose GT). The online map is a tracking
reference rather than a reconstruction (10--16~dB);
after the offline pose--map refinement and global reconstruction, our renders recover
high-frequency appearance detail (23--38~dB) and lead Gaussian-LIC2
(14--34~dB) on every selected view.}
\label{fig:qualitative}
\end{figure}

Figure~\ref{fig:qualitative} visualizes this margin on four sequences.
The upstream map blurs signage and fine structure, whereas our global reconstruction restores it.
Table~\ref{tab:lic2} shows the comparison, which can show that our method led on all six sequences by 5.4 to 12.3~dB.
The self-reported values of Gaussian-LIC2 \cite{lang2025gaussianlic2} agree with our values within 0.2~dB.

These results between refined 3DGS and our online map and Gaussian-LIC2 show that the online map is a tracking reference rather than a reconstruction.
The offline pose--map refinement and global reconstruction can recover high-frequency appearance detail.

\begin{table}[t]
\centering
\caption{\textbf{Pose--map decomposition ablation on the 237-keyframe archive under an equal budget.}
Map-only freezes the keyframe poses at the SLAM estimates and only trains the Gaussian map;
pose-only freezes the seed map and only refines the poses.}
\small
\begin{tabular}{@{}l c c c@{}}
\toprule
Configuration & Optimized terms & $\mathrm{PSNR}$ (dB) & $\Delta$ (dB) \\
\midrule
Alternating (full) & poses $+$ map & 30.06 & --- \\
Map-only & map only & 27.01 & $-3.05$ \\
Pose-only & poses only & 6.64 & $-23.42$ \\
\bottomrule
\end{tabular}
\label{tab:posmap}
\end{table}

We also conducted an ablation study on the pose--map decomposition to further justify the alternating design of Eq.~\eqref{eq:base}.
The same archived keyframes are replayed under an equal budget with one optimized term ablated each time.
As Table~\ref{tab:posmap} shows, alternating pose--map optimization reaches 30.06~dB, map-only training from frozen poses loses 3.05~dB (27.01~dB), and pose-only refinement over a frozen seed map collapses to 6.64~dB.
It shows that both optimized terms (poses and Gaussians) are necessary for this method.

\subsection{Experiment 3: Relocalization Evaluation}
\label{sec:exp-reloc}
This experiment evaluates the accuracy of the relocalization module in recovering poses from the final generated 3DGS map using the views unseen during mapping.
We also evaluate the reliability in inheriting the extrinsic quality of that map.
All query sets exclude map-building frames.
The map is built from keyframes sampled at a fixed stride, and the remaining frames serve as queries for this relocalization experiment.
For our self-collected sequence, an \emph{A/B relocalization experiment} is also conducted to compare two groups by injecting the $0.153^\circ$ and $1.51^\circ$ residual to the calibrated configuration, to show the effect of extrinsic error on relocalization performance.
Here, only the extrinsic is different and other configuration parameters are identical: the same bag with the same XFeat engine, the same camera intrinsics, and the same 100-frame query.

\begin{table}[t]
\centering
\caption{\textbf{Result of held-out relocalization (117 queries for the self-collected sequence, 100 for the others).}
} 
\footnotesize
\setlength{\tabcolsep}{3.2pt}
\begin{tabular}{@{}l l c c c c c c@{}}
\toprule
 & & \multicolumn{4}{c}{Image-based route} & \multicolumn{2}{c}{Point-cloud route} \\
\cmidrule(lr){3-6}\cmidrule(l){7-8}
Sequence & Resolution & Success & Inlier ratio & RMSE (px) & $\mathrm{PSNR}$ (dB) & ICP success & $t_{\mathrm{med}}$/$r_{\mathrm{med}}$ \\
\midrule
Ours A, stride-2 map & 1280$\times$720 & 100\% (117/117) & 0.647 & 2.30 & 29.66 & 100/100 & 1.6 / 0.28 \\
Ours B, stride-2 map & 1280$\times$720 & 100\% (117/117) & 0.592 & 2.36 & 28.64 & --- & --- \\
tuhh\_day\_04 & 640$\times$480 & 100\% (100/100) & 0.517 & 2.37 & 27.68 & 99/100 & 1.6 / 0.24 \\
hku\_campus & 1280$\times$1024 & 95\% (95/100) & 0.285 & 2.61 & 29.52 & 100/100 & 1.7 / 0.23 \\
Retail & 640$\times$512 & 86\% (86/100) & 0.210 & 2.27 & 29.42 & 99/100 & 8.2 / 1.23 \\
CBD & 1280$\times$1024 & 35\% (35/100) & 0.216 & 2.63 & 30.01 & 100/100 & 6.4 / 0.82 \\
Bright & 1280$\times$1024 & 0\% (0/100) & --- & --- & --- & 98/100 & 8.2 / 1.18 \\
\bottomrule
\end{tabular}
\label{tab:heldout}
\end{table}

Table~\ref{tab:heldout} shows the main held-out results.
In normal-texture scenes, the success rate is entirely insensitive to the held-out protocol and remains 100\% even after excluding the map-building frames.
Two sequences (hku\_campus and Retail) are mildly sensitive, changing from 96\% to 95\% and from 91\% to 86\%, but the success rates still remain high and can be considered to be acceptable.
For CBD, even in the weak-texture glass-curtain-wall scene, our method can still achieve 35\% success rate.
However, for the extreme weak-texture scene Bright, the relocalization module fails to recover any pose.

The result shows that the relocalization module is robust to view gaps and can recover accurate poses from the final generated 3DGS map.
However, for extreme weak-texture scenes, the visual route may fail to recover accurate poses, due to the lack of sufficient features for matching and pose estimation. 
The geometric ICP route runs after the visual route, which can cover this failure mode to a large extent.
On CBD and Bright, where the visual success drops to 35\% and 0\%, it still succeeds on 100/100 and 98/100 queries, with median translation and rotation errors of 6.4~cm/0.82$^\circ$ and 8.2~cm/1.18$^\circ$.
The two routes are complementary: the visual route delivers pixel-level accuracy in normal-texture scenes, while the geometric route keeps the pose recoverable from map geometry alone where appearance cues fail.

\begin{table}[t]
\centering
\caption{\textbf{Result of relocalization for self-collected sequence with different extrinsic errors.}}
\small
\begin{tabular}{@{}l c c p{2.0cm} c c@{}}
\toprule
Configuration & Extrinsic residual & Inlier ratio & Low-inlier frames ($<0.65$) & Rendered $\mathrm{PSNR}$ & RMSE (px) \\
\midrule
EXTROT-A & 0.153$^\circ$ & 72.3\% & 2/100 & 30.09~dB & 2.15 \\
EXTROT-B & 1.51$^\circ$ injected & 68.0\% & 34/100 & 29.15~dB & 2.21 \\
\bottomrule
\end{tabular}
\label{tab:ab}
\end{table}

Table~\ref{tab:ab} shows the paired A/B comparison over 100 frames, with the different extrinsic residuals of 0.153$^\circ$ and 1.51$^\circ$.
The inlier-ratio degradation is 4.4 percentage points, with paired $|t| \approx 9.3$ and Wilcoxon $z \approx 8.7$ at $p < 0.001$; the rendered $\mathrm{PSNR}$ degrades by 0.94~dB; and 86\% of frames degrade in the same direction, with the paired median difference at 0.88~dB.
Both groups succeed on all frames, which shows that a 1.51$^\circ$ error lies within the robust bound since RANSAC with mutual-nearest filtering absorbs the wrong matches.
This result shows that our method is robust enough for the 3DGS model even for different quality levels.

\section{Conclusion}
\label{sec:concl}
In this paper, we presented a novel method---GEMOR-GS, which is a LiDAR--inertial--visual SLAM system that upgraded the 3DGS map from an online render asset to a dependable robot world model.
A direct-method calibration module estimated the LiDAR--camera extrinsic offline over dedicated calibration segments and supplied it to both the online and the offline tracks.
The offline track refined keyframe poses and the Gaussian map alternately, purifying the depth--color supervision that drove Gaussian training.
A global reconstruction stage re-initialized the Gaussian model from the refined keyframe schema with all online constraints lifted, executing global densification, pruning, and spherical-harmonics order elevation.
Finally, the relocalization module matched XFeat descriptors against a lightweight relocalization database, obtained a coarse pose by RANSAC-PnP, and refined it by Gaussian RGB-D warp alignment, enabling the robot to recover its global pose after tracking loss.
In the experiments, we validated the effectiveness of each module.
The direct-method calibration can recover injected extrinsic offsets with a small error (rotation outputs of 0.025--0.227$^\circ$ and translation outputs of 4.3--39.9~mm). 
The calibration-refined global reconstruction won all 6,632 paired keyframes against the online map by 14.3 to 19.9~dB, led same-bag Gaussian-LIC2 by 9.43~dB on the zero-protocol-difference sequence.
The relocalization module recovered poses from the final 3DGS map with held-out success rates on most sequences, and its geometric ICP route kept the pose recoverable even in the extreme weak-texture scene where visual matching collapsed.
These results show the effectiveness of the proposed method.
As a Gaussian world model, the resulting map is geometrically accurate and appearance-faithful enough to be queried for perception and planning, and it is re-enterable enough for robots to rely on over continuous operation.
\ifPAIanonymous
\else
The source code of our method is publicly available at \url{https://github.com/Vector-4D/gemor-gs}.
\fi
Future work includes feeding the recovered pose back into the online filter to resume mapping after tracking loss, generative completion of unobserved viewpoints.
To use it on real robots, we also need to improve the conversion of the Gaussian map to mesh and collision geometry for planning and manipulation, semantic integration with object-level structure.

\subsection*{Review-stage declarations}
\textbf{Competing interests:} The authors declare no competing interests.
\textbf{Data and materials availability:} The public datasets used are available from their original sources.
\ifPAIanonymous
\else
The source code of GEMOR-GS is publicly available at \url{https://github.com/Vector-4D/gemor-gs}.
Source code and the self-collected dataset will be made publicly available upon publication.
\fi

\bibliographystyle{PAI}
\bibliography{GEMORGS_references}

\end{document}

%% file: PAI_author_details.tex
\author{
Ming~Ding$^{1\ast}$,
Yuchen~Yang$^{1}$,
Xinyu~Wu$^{1}$,
Wei~Wang$^{1}$,
Hesheng~Wang$^{1}$\and
\small$^{1}$Nanjing 4D Vector Technology Co., Ltd., Nanjing, China.\and
\small$^\ast$Corresponding author. Email: ming@vector4d.com
}